# 2D Self-Organized ONN Model For Handwritten Text Recognition

Hanadi Hassen Mohammed[1], Junaid Malik[2], Somaya Al-Madeed[1], and Serkan Kiranyaz[3], Senior Member, IEEE

[1]Department of Computer Science and Engineering, Qatar University, Doha , Qatar (e-mail: hm1409611@qu.edu.qa, s_alali@qu.edu.qa).
[2]Department of Computing Sciences, Tampere University, Tampere, Finland. (e-mail: junaid.malik@tuni.fi).
[3]Department of Electrical Engineering, Qatar University, Doha, Qatar (e-mail: mkiranyaz@qu.edu.qa).

Abstract— Deep Convolutional Neural Networks (CNNs) have recently reached state-of-the-art Handwritten Text Recognition (HTR) performance. However, recent research has shown that typical CNNs' learning performance is limited since they are homogeneous networks with a simple (linear) neuron model. With their heterogeneous network structure incorporating non-linear neurons, Operational Neural Networks (ONNs) have recently been proposed to address this drawback. Self-ONNs are self-organized variations of ONNs with the generative neuron model that can generate any non-linear function using the Taylor approximation. In this study, in order to improve the state-of-the-art performance level in HTR, the 2D Self-organized ONNs (Self-ONNs) in the core of a novel network model are proposed. Moreover, deformable convolutions, which have recently been demonstrated to tackle variations in the writing styles better, are utilized in this study. The results over the IAM English dataset and HADARA80P Arabic dataset show that the proposed model with the operational layers of Self-ONNs significantly improves Character Error Rate (CER) and Word Error Rate (WER). Compared with its counterpart CNNs, Self-ONNs reduce CER and WER by 1.2% and 3.4 % in the HADARA80P and 0.199% and 1.244% in the IAM dataset. The results over the benchmark IAM demonstrate that the proposed model with the operational layers of Self-ONNs outperforms recent deep CNN models by a significant margin while the use of Self-ONNs with deformable convolutions demonstrates exceptional results.

Index Terms— Handwriting Text Recognition, Self-organized Operational Neural Networks, Generative Neurons, Deformable Convolution

## 1 Introduction

OFFLINE Handwritten Text Recognition (HTR) is the process of digitizing text that is shown in a picture. It is a well-known topic in the field of Computer Vision; however, it is regarded as a challenging task for many reasons such as the variation and ambiguity of strokes from person to person. In addition, a person's handwriting style can vary from time to time and the degradation of the source document image over time causes a loss of characters. These challenges make it difficult to create an effective, generalizable system.

Recurrent Neural Networks (RNNs), and Multidimensional Long Short-Term Memory (MDLSTM) networks, have been extensively used in HTR. MDLSTM networks show state-of-the-art performance on most of the HTR benchmarks. Regular LSTM networks differ from MDLSTM in that the former introduces recurrence along the axis of one-dimensional sequences, whereas the latter introduces recurrence along two axes which makes it ideal to handle unrestricted two-dimensional input. In line-level HTR, it is common to use the MDLSTM to extract features. The character-level transcription of the input line image is then obtained by converting the 2D data into a 1D sequence. This design is fundamentally at the heart of most of the successful line-level HTR techniques; however, compared to Convolutional Neural Networks (CNNs), MDLSTMs are computationally expensive. Furthermore, a visual comparison of the 2D-LSTM features retrieved in the bottom layers reveals that they are visually similar to the 2D CNN outputs [1].

CNNs are well known for feature representation of input images. Recently, they have been adopted in handwriting recognition models in combination with Recurrent Neural Networks (RNNs) which are responsible for generating output sequences and decoding the underlying text [2]. In [3], three CNN layers are used to extract features from input images which are then fed into two CNN-MDLSTMs for extracting context information. The proposed model in [4] incorporates CNNs with MDLSTMs, but instead of setting CNNs as feature extractors for the input images, layers of LSTMs scan the blocks of input images in different directions, then, CNN layers receive the output of each LSTM layer and again forward to LSTMs. The top-most layer is fully connected rather than convolutional. The softmax layer receives the last activations, which are summed vertically. Connectionist Temporal Classification (CTC) is used to process softmax's output. A similar idea is proposed in [5], but they propose an optimized version of MDLSTM where the convolutional and recurrent layers have been relocated, and the subsampling processes have been tweaked to improve feature extraction at the bottom levels while lowering the activation volume before reaching the upper layers. An HMM is used in the decoding step to reduce errors generated by the CNN- MDLSTM optical model.

A recent study [6] introduced a convolutional-only architecture for HTR. They use deformable convolutions [7] to tackle the problem of diversity in writing styles as the deformation of the kernel can be interpreted as geometrical deformations of the same textual elements. Deformable convolutions broaden the definition of convolution by redefining the shape of the convolution as adaptable. Convolution weights are supposed to multiply inputs not on the conventional orthogonal, canonical k × k grid, but rather on a learning-based weight-input coordinate correspondence. The state-of-the-art performance level was achieved in [6] by reducing character uncertainty at the network's softmax output.

According to recent studies, [8]–[11], CNNs, like its predecessors, Multi-Layer Perceptrons (MLPs), rely on the ancient linear neuron model, so they are successful in learning linearly separable problems very well, but they may completely fail when the problem's solution space is highly nonlinear and complex. Operational Neural Networks (ONNs) [12], are recently proposed heterogeneous networks with a non-linear neuron model. They can learn highly complex and multi-modal functions or spaces even with compact architectures. Similar to Generalized Operational Perceptrons (GOPs) [13], [14], operational neurons of ONNs are modeled similar to biological neurons, with nodal (synaptic connections) and pool (synaptic integration in the soma) operators. An operator set is a collection of the nodal, pool, and activation operators, and the operator set library needs to be built in advance to contain all possible operator sets. ONNs, too, have a variety of limits and downsides as a result of such fixed and static architecture. First, only the operators in the operator set library can obviously be used, and if the correct operator set for the learning problem at hand is not in the library, the desired learning performance cannot be achieved. Second, to reduce the search space one or few operator sets can be assigned to all neurons in each hidden layer, which poses a limited level of heterogeneity. Finally, there is always a need for searching for the best operators sets for each layer, which might be cumbersome, especially for deeper networks.

To tackle the aforementioned problems, the authors in[15] proposed self-organized operational neural networks (Self-ONNs) with generative neurons. The generative neuron model allows Self-ONNs to self-organize by iteratively generating nodal operators during the back-propagation (BP) training to maximize the learning performance. Certainly, being able to create any non-linear nodal operator significantly improves both operational diversity and flexibility.

Self-ONNs are the super-set of the conventional CNNs. Contrary to the CNNs' homogenous network structure with only the linear neuron model, Self-ONNs are heterogenous networks with a "self-generating" non-linear neuron model (i.e., the generative neurons). This yields superior diversity and learning performance. Thus, Self-ONNs in numerous problems such as severe image restoration [11], R-peak detection in low-quality Holter ECGs [8], patient-specific ECG classification [9] and biometric authentication [16] outperformed their equivalent and even deep CNNs with a significant performance gap.

The significant and novel contributions of this study are: 1) The pioneer application of Self-ONNs in HTR is proposed. 2) An adequate level of non-linearity of the operational layers to boost the recognition performance under various topologies is investigated. 3) A further investigation is carried out on the use of deformable convolutions along with the generative neurons in the same network . 4) The state-of-the-art performance is achieved with a significant gap against the recent methods on the IAM English dataset while the performance gap is further widened on the HADARA80P Arabic dataset

The rest of this paper is organized as follows: Section 2 summarizes the main features of Self-ONNs with generative neurons. Section 3 describes the methods used in this paper, the proposed architecture, datasets used, and implementation details of the proposed HTR system. Section 4 presents the experimental results including evaluation of the proposed system, the statistical analysis, and a detailed set of comparative evaluations against recent methods. Section 5 concludes the paper and discusses future research directions.

## 2 SELF-ORGANIZED OPERATIONAL NEURAL NETWORKS

For Self-ONNs, the generative neuron model was proposed recently to overcome the limitations of ONNs, in particular, the requirement for a pre-defined set of operators and the need for searching inside that library to discover the best set of operators. Self-ONNs promise a similar or better performance level than conventional ONNs with an elegant diversity and reduced computational complexity.

Let us define the input tensor to a layer by $X \in R^{H \times W \times C_{in}}$ and a sub-tensor of it centered at the position $i, j$

by $X_{(i,j)} \in R^{h \times w \times C_{in}}$. Let us define $W_{c_{out}} \in R^{h \times w \times C_{in}}$, $c_{out} = 1, \ldots, C_{out}$ the $c_{out}'$th filter of a layer. In CNNs, convolutional neurons convolve $X$ with $W_{c_{out}}$ and add an offset, which equates to doing the following calculation for each point $(i, j)$ of the input tensor $X$:

$$Y_{c_{out}}(i, j) = \sum_{k,m,c_{in}=1}^{h,w,C_{in}} W_{c_{out}}(k, m, c_{in}) X_{(i,j)}(k, m, c_{in}) + b_{c_{out}}$$
$$= w_{c_{out}}^T x_{(i,j)} + b_{c_{out}} \qquad (1)$$

the $Y_{c_{out}}(i, j)$ is the $(i, j)'^{th}$ element of the output feature map $Y_{c_{out}}$, $b_{c_{out}}$ is the bias term, and $x_{(i,j)}$ and $w_{c_{out}}$ are vectorized versions of $X_{(i,j)}$ and $W_{c_{out}}$, respectively. The feature mappings $Y_{c_{out}}$, $c_{out} = 1, \ldots, C_{out}$ are concatenated to generate the tensor $Y \in R^{H \times W \times C_{out}}$, and the layer output is produced using an element-wise activation function. ONNs generalize CNNs transformation in (1) using:

$$Y_{c_{out}}(i, j) = \Psi(x_{(i,j)}, w_{c_{out}}) + b_{c_{out}} \qquad (2)$$

where $\Psi$ is a nodal function that can be a combination of different functions. During the training, the selection of $\Psi$ is achieved using a search strategy. The ONN layer is a conventional CNN layer when the nodal function is determined to be the dot-product of its arguments. The operator chosen will be applied to every connection and every kernel element in the network [17].

Instead of searching for the best possible nodal function, during training, each generative neuron in a Self-ONN can iteratively generate any nonlinear function of each kernel element with a truncated Mac-Laurin series expansion:

$$\Psi(x_{(i,j)}, w_{c_{out},1},\ldots,w_{c_{out},Q}) = w_{c_{out},1}^T x_{(i,j)} + w_{c_{out},2}^T x_{(i,j)}^2 + \cdots + w_{c_{out},Q}^T x_{(i,j)}^Q = \sum_{q=0}^{Q} w_{c_{out},q}^T x_{(i,j)}^q, \qquad (3)$$

where $x_{(i,j)}^q$ is an element-wise power, $w_{c_{out},q}$ are learnable weights interacting with $x_{(i,j)}^q$. As a result, each neuron undergoes the following transformation:

$$Y_{c_{out}}(i, j) = \sum_{q=1}^{Q} w_{c_{out},q}^T x_{(i,j)}^q + b_{c_{out}}, \qquad (4)$$

where $w_{c_{out},q}$, $q = 1, \ldots, Q$, are learned using gradient-based optimization since the purpose is to learn the best suited nodal function. For more detailed information and details on the BP formulations, the readers are referred to [15].

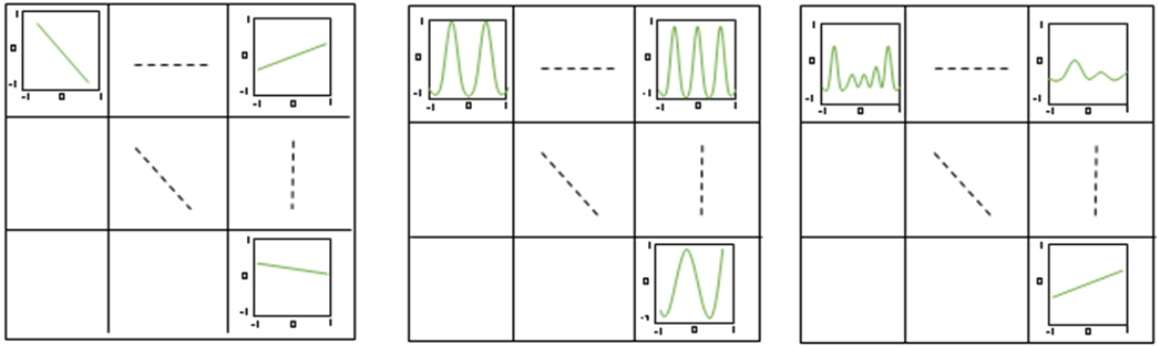

Fig. 1. Kernels in CNNs (left), ONNs (middle) and Self-ONNs (right).

## 3 METHODS

To better understand the difference between CNNs, ONNs, and Self-ONNs, Fig.1 illustrates how the kernels may look like in these three networks. In conventional CNNs (Fig. 1. (left)), a linear transformation is always used in convolution with the input tensor. In ONNs (Fig. 1. (middle)), a selected non-linear operator is used for all kernel elements (e.g. sinusoids with different frequencies). In Self-ONNs (Fig. 1. (right)) the right nodal operator for every kernel element, every neuron, and every synaptic connection is generated during (BP) training. This allows that in Self-ONNs, for a certain kernel element the nodal operator can be linear while for another may be similar to a sinusoid or any arbitrary non-linear function. This allows not only neuron-level but even kernel-level diversity and heterogeneity.

To investigate the impact of using a heterogeneous and non-linear network model in HTR, certain modifications are made to the recently proposed CNN-only HTR system [6] that currently holds the state-of-the-art HTR performance. This section goes over the modifications made to the blocks and some additional system features.

### 3.1 The Proposed Architecture

As shown in Fig. 2 (a), the proposed architecture consists of two parts; the backbone and the head. Fig.2 (b) shows how the deformable convolutional are inserted in the model. The backbone consists of a group of ResnetBlocks [18] blocks and acts as an optical model responsible for transforming the input images into feature maps. Each block contains either 2D-CNNs or 2D-Self-ONNs with 3 × 3 kernels, 1 × 1 stride, 1 × 1 padding, and 1 × 1 dilation. The number of filters in each group of blocks is twice the number of filters in the previous group of blocks. Each CNN or Self-ONN layer is followed by batch normalization and each group of blocks is followed by max pooling.

The feature maps extracted from the convolutional backbone are then fed into the convolutional or operational head to be transformed into character predictions with the help of either 1D-CNN or 1D-Self-ONN. The convolutional or operational head consists of several CNNs or Self-ONNs, each one is followed by batch normalization and a ReLU non-linearity (in the case of CNN) or Tanh (in the case of Self-ONN). The sequence of probability distributions over the potential characters is then generated using the softmax function on the final output, which is then propagated into a Connectionist Temporal Classification (CTC) loss [19].

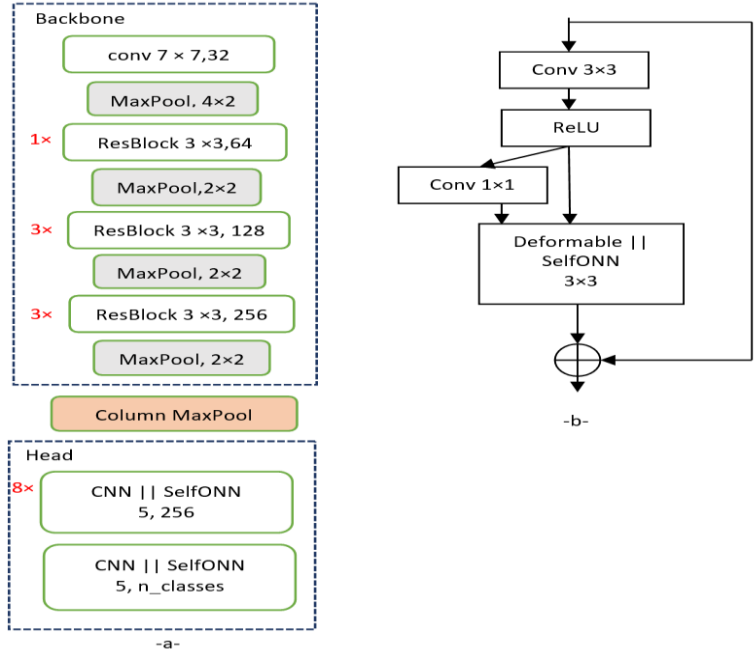

Fig. 2. (a) System architecture consisted of a Backbone that contains either CNNs or Self-ONNs, and a Head (1D-CNNs or 1D-Self-ONNs). (b) ResBlock variation for Self-ONNs or deformable convolutions.

For an input $(x)$ of length $(T)$, $x = (x_1, \ldots, x_T)$ and output $(y)$ of length $(U)$, $y = (y_1, \ldots, y_U)$, where $y_u = \in \{1, \ldots, K\}$, K is the number of target labels, the main idea of CTC is to align the prediction to the target using an intermediate label representation $\pi = (1, \ldots, \pi_T)$. CTC allows for label repetitions and the appearance of a blank label (–), which symbolizes the special emission without labels. During CTC training, the model maximizes $P(y|x)$ across all possible $\Phi(y')$ label sequences, where $y'$ is a modified label sequence of $y$:

$$P(y|x) = \sum_{\pi \epsilon \Phi(y')} P(\pi|x) \tag{5}$$

The $y'$ allows blanks in the output by inserting blank symbols between each label and the beginning and end e.g.(y = (h,e,n), $y' = (-, h, -, e, -, n, -)$). The label sequence probability $P(y|x)$ is calculated using:

$$P(y|x) = \prod_{t=1}^{T} q_t(\pi_t), \tag{6}$$

where $q_t(\pi_t)$ represents the softmax activation of the $\pi_t$ label in the network output layer q at time t. The CTC loss to be minimized is defined as the negative log-likelihood of the ground truth character sequence $y^*$:

$$\mathcal{L}_{CTC} \triangleq -\ln P(y^*|x) \tag{7}$$

Using the forward-backward algorithm, the $P(y^*|x)$ is computed as:

$$P(y^*|x) = \sum_{u=1}^{|y'|} \frac{\alpha_t(u)\beta_t(u)}{q_t(y'_u)}, \tag{8}$$

where $\alpha_t(u)\ and\ \beta_t(u)$ are the forward and backward variables respectively. The forward variable $\alpha_t(u)$ represents the total probability of all possible prefixes $(y'_{1:u})$ ending with the u-th label while the backward variable $\beta_t(u)$ represents all possible suffixes$(y'_{u:U})$ starting with the u-th label. Then, back-propagation can be used to train the network by taking the derivative of the loss function with respect to $q_t(k)$ for any k label, including the blank. The details of the forward-backward algorithm can be found in [19].

### 3.2 Datasets

To test the proposed approach, a widely used line-level dataset IAM [20] containing 9,862 text lines is used. It includes a training set consisting of 6161 lines, a testing set consisting of 1861 lines, and two validation sets consisting of 900 and 940 lines.

Another benchmark dataset, HADARA80P [21], which is based on an Arabic historical handwritten book, is used. It contains 1,336 lines. In this research, 80% of the dataset for training, 10% for validation, and 10% for testing were used. Fig. 3 shows samples for both datasets. The HADARA80P dataset has some segmentation problems as illustrated in Fig 3. (b) with green circles. In addition, some dots are missed as illustrated with the red circle. The dots are essential in Arabic writing, some letters have the same shape and only can be differentiated through dots above or below a letter e.g. letters (ب، ت، ث). Because historical documents are prone to degradation, many characters lose their dotting which makes the problem of automatic recognition more challenging in Arabic.

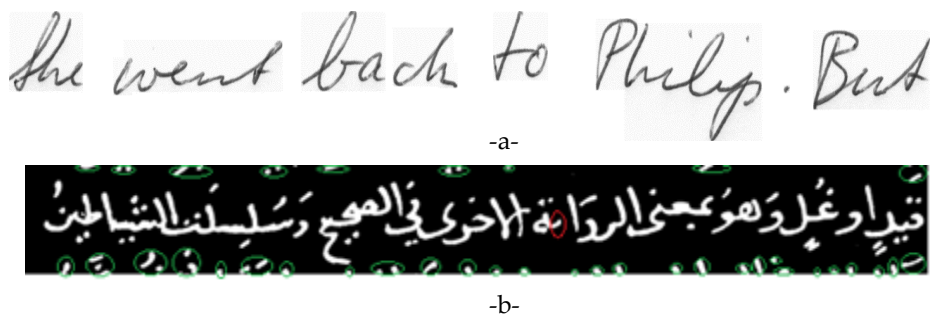

-a-

-b-

Fig. 3. (a) Image from the IAM dataset. (b) Image from the HADARA80P dataset. The red circle highlights a character with missing dots. The green circles highligh segmentation errors.

### 3.3 Implementation Details

Because the Hadara80P dataset is much smaller than the IAM dataset, only 120 epochs were trained, while for the IAM dataset, the number of epochs was increased to 2000 epochs. Adam optimizer was used to train the model with a maximum learning rate of 4e−5 (for IAM) and 1e-5 (for Hadara80P) while the batch size was set to 12 for both.

### 3.4 Evaluation Metrics

As in the prior studies in HTR, the Word Error Rate (WER) and the Character Error Rate (CER) are the common evaluation metrics used in this study, both of which use the Levenshtein Distance [22] between the predicted text and the target text. The ratio of misrecognized characters is represented by CER, whereas the ratio of misrecognized words is represented by WER.

Three types of errors need to be considered when calculating the CER and WER: the substitution error (S) is the misspelled characters or words, the deletion error (D) is the lost or missing characters or words, and the insertion error (I) is the incorrect inclusion of characters or words. The following formula describes the common calculation of CER:

$$CER = \frac{S+D+I}{N}, \qquad (9)$$

where N is the number of characters in the ground truth. The WER formula is similar to CER, but at the word level.

$$WER = \frac{S_w + W_w + I_w}{N_w} \qquad (10)$$

If the Levenshtein distance between two words is not zero, the word is considered incorrectly classified even if only one character is incorrect. This evaluation includes all symbols including special characters. Fig. 5 shows samples taken during the testing of the proposed model along with CER and WER.

## 4 EXPERIMENTAL RESULTS

This section presents the performance of replacing the CNN layers with SelfONN layers in the original model and the effect of changing the non-linearity level (Q orders) on the model performance. Then the performance of the modified architecture is analyzed, and the comparison of the proposed architecture with other works that use the same dataset is presented.

### 4.1 Evaluating The Self-ONN Optical Model

In the first experiment, the performance of CNNs versus Self-ONNs is compared using the original architecture proposed in [6] which consists of 10 blocks in the backbone and three convolutional layers in the head. Table 1 illustrates the comparison of accuracy after replacing the CNN layers in the head with operational layers with Q = 3, 5, 7, 9 (the order of the Taylor polynomials). The performance of the models with the best CER and the WER is reported. The results show that Self-ONNs consistently outperform both CNNs and deformable CNNs in terms of CER and WER.

The model is further modified by removing three blocks from the convolutional backbone leading to a reduction of 6 CNN layers. This compact architecture is tested on the HADARA80P dataset because it is smaller than the IAM dataset. The results in Table 2 show that using operational layers on the backbone was yielding superior results than using them on the head. The results also show an exceptional accuracy improvement (3.464 % and 1.2 % on WER and CER respectively) when using SelfONNs compared to the native CNN-only architecture.

TABLE 1. COMPARISON OF THE PERFORMANCE OF CNNS, SELF-ONNS AND DEFORMABLE CONVOLUTIONS ON IAM DATASET USING THE ORIGINAL ARCHITECTURE PROPOSED IN [6]

| Configuration | | Q-order | Best CER | | Best WER | |
|---|---|---|---|---|---|---|
| Backbone | Head | | CER | WER | CER | WER |
| CNN | CNN | - | 5.424 | 18.914 | 5.171 | 17.894 |
| CNN | Self-ONN | 3 | 5.128 | 17.982 | 5.128 | 17.982 |
| CNN | Self-ONN | 5 | 5.145 | 17.858 | 5.145 | 17.858 |
| CNN | Self-ONN | 7 | 5.202 | 18.178 | 5.202 | 18.178 |
| CNN | Self-ONN | 9 | 5.270 | 18.258 | 5.270 | 18.258 |
| **CNN** | **Self-ONN** | **3,5,7** | **5.075** | **17.589** | **5.075** | **17.589** |
| CNN+DeformableCNN | Self-ONN | 3,5,7 | 5.156 | 17.771 | 5.151 | 17.654 |

TABLE 2. COMPARISON OF THE PERFORMANCE OF CNNS, SELF-ONNS (Q = 3,5,7 IN ALL), AND DEFORMABLE CONVOLUTIONS ON THE HADARA80P DATASET USING ONLY 7 BLOCKS ON THE BACKBONE.

| Configuration | | Best CER | | Best WER | |
|---|---|---|---|---|---|
| Backbone | Head | CER | WER | CER | WER |
| CNN | CNN | 9.199 | 35.912 | 9.191 | 35.460 |
| CNN+DeformableCNN | CNN | 9.639 | 37.216 | 9.639 | 37.216 |
| CNN | Self-ONN | 12.486 | 46.178 | 12.486 | 46.178 |
| **Self-ONN** | **CNN** | **7.977** | **32.038** | **7.991** | **31.996** |
| Self-ONN+DeformableCNN | CNN | 11.764 | 46.010 | 12.145 | 46.681 |

The model is further improved for the IAM dataset by adding the removed layers to the head leading to a total of 9 CNN layers in the head. The results are presented in Table 3. This new model where layers in the feature extraction part (backbone) are reduced and layers in the classification part (head) are populated leads to a better CER or WER than the ones reported in [6]. Following this, using operational layers of Self-ONNs in the feature extraction part exhibits even better results than using only CNNs or CNNs with deformable convolutions. Finally, the combination of Self-ONNs with deformable convolutions achieves an exceptional improvement in both CER and WER and thus setting a new state-of-the-art performance level in this domain.

An important observation worth mentioning is that although the use of the three neuron types (convolutional, deformable convolutional, operational) improved the performance, the location of each one in the network architecture also had an important effect. By looking at the results in Table 1, although the three networks were used (CNN and deformable in the backbone while Self-ONN in the head), the result was not the best. Based on the best results achieved on the IAM and the HADARA80P datasets, the operational layers were better performing especially when used at the beginning of the network (closer to the input layer). Usually, the first few layers in CNNs capture the low-level features, whereas the following layers extract high-level. As a handwritten image primarily consists of straight and curved lines rather than objects, having superior features in the initial layers of the model is essential. Therefore, in this particular HTR problem, it is recommended to use operational layers at the beginning of the network.

To analyze the complexity of the proposed model, the number of trainable parameters is reported in Table 4. The Self-ONNs-based models are adding more complexity however this complexity can be tackled with the use of GPUs.

TABLE 3. COMPARISON OF THE PERFORMANCE OF CNNS, SELF-ONNS (Q = 3,5,7 IN ALL), AND DEFORMABLE CONVOLUTIONS ON THE IAM DATASET USING THE PROPOSED ARCHITECTURE IN FIG. 2.

| Configuration | | Best CER | | Best WER | |
| --- | --- | --- | --- | --- | --- |
| Backbone | Head | CER | WER | CER | WER |
| CNN | CNN | 4.799 | 16.287 | 4.731 | 16.324 |
| CNN+DeformableCNN | CNN | 4.914 | 16.171 | 4.914 | 16.171 |
| CNN | Self-ONN | 4.737 | 16.164 | 4.895 | 16.2583 |
| Self-ONN | CNN | 4.732 | 16.033 | 4.794 | 16.127 |
| **Self-ONN+DeformableCNN** | **CNN** | **4.576** | **15.488** | **4.532** | **15.080** |

TABLE 4. COMPARISON OF MODEL COMPLEXITY IN TERMS OF NUMBER PARAMETERS

| Configuration | Trainable Parameters |
| --- | --- |
| CNN | 6,890,768 |
| CNN + Deformable | 6,912,782 |
| Self-ONN | 29,943,056 |
| CNN&DeformableCNN&Self-ONNs | 17,505,038 |

### 4.2 Statistical Analysis

The WER and CER per image are used to conduct the statistical test to verify the significance of the proposed models. Fig.4. shows the boxplots of the generated errors by the best-performing model and its counterpart CNN model in both datasets. The Wilcoxon test is used to see if there is a significant difference in average between two paired samples. The Wilcoxon test here is used to evaluate if there is any significant reduction in error rates after using operational layers. The results of this test are shown in Table 5 and Table 6. All the p-values are below 0.05 which indicates the significant difference in prediction errors generated by the CNN-based model and the Self-ONN-based model in both datasets.

A close look to Fig. 4 will reveal the fact that the performance gap between CNN and Self-ONN models widens in the HADARA80P dataset. This is an expectable outcome since the HADARA80P dataset is more noisy and contains fewer data compared to the IAM dataset. This indicates the superiority of the Self-ONN models in working with degraded or highly noised manuscripts.

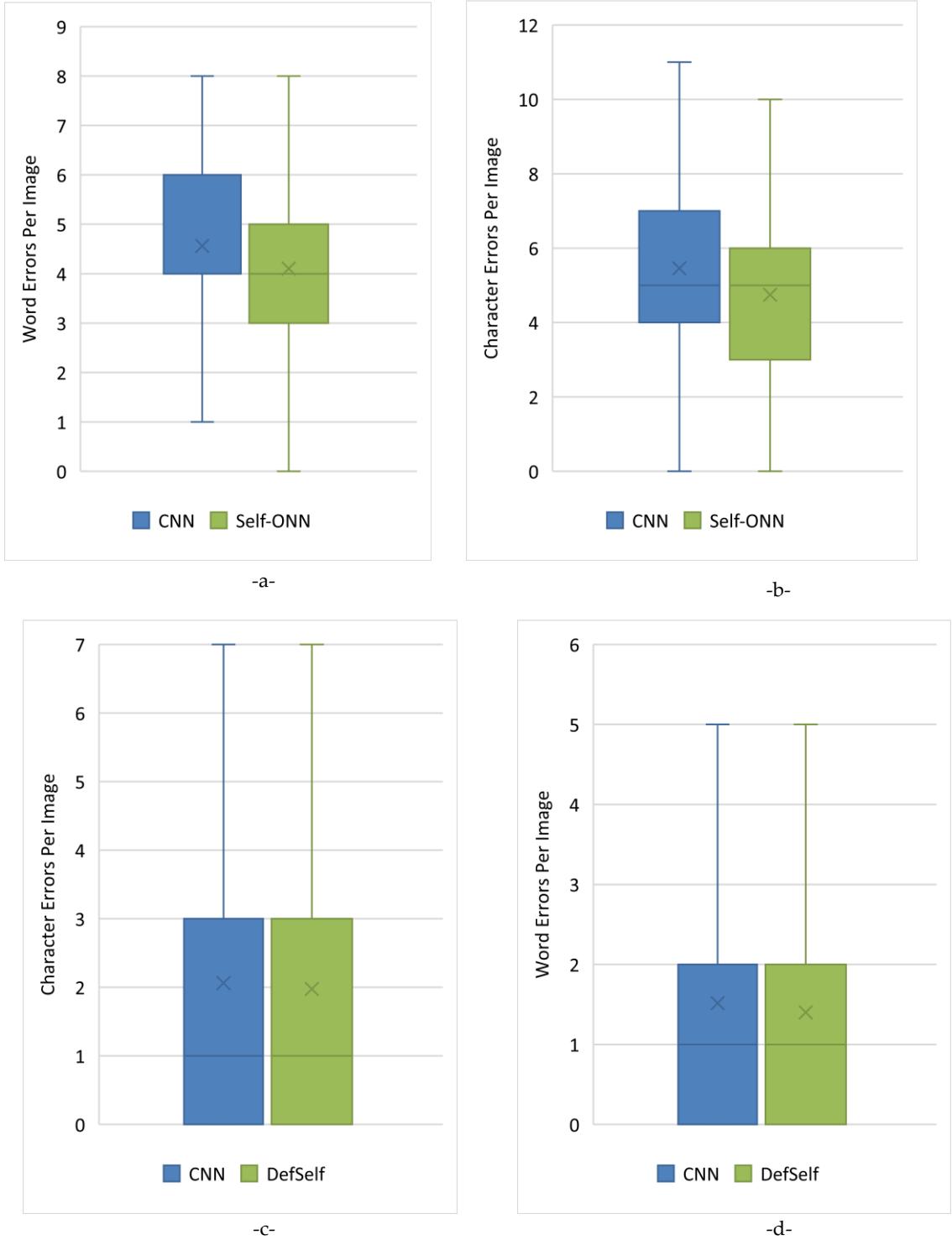

Fig. 4. WER and CER Box plots for HADARA80P dataset (a and b) and IAM dataset

TABLE 5. WILCOXON SIGNED RANKS TEST FOR HADARA80P DATASET RESULTS

|  | CNN_CERs - SelfONN_CERs | CNN_WER - SelfONN_WER |
|---|---|---|
| Z | -16.083 | -13.239 |
| Asymp. Sig. (2-tailed) | .000 | .000 |

TABLE 6. WILCOXON SIGNED RANKS TEST FOR IAM DATASET RESULTS

|  | CNN_CERs - DefSelf_CERs | CNN_WERs - DefSelf_WERs |
|---|---|---|
| Z | -2.154[b] | -4.153[b] |
| Asymp. Sig. (2-tailed) | .031 | .000 |

## 4.3 Performance Evaluations

As discussed earlier, several deep learning models of HTR were proposed in the literature including LSTM-Based approaches, attention-based approaches, and sequence-to-sequence transition approaches, in addition to the recently proposed CNN-only approaches. For a detailed set of comparative evaluations, the performances of the proposed methods are compared against all recent state-of-the-art methods (lexicon-free in the line-level) and presented in Table 7. The proposed method outperforms all prior works. The proposed Self-ONN architecture combined with deformable convolution improved the CER by 0.14% and WER by 1.49% over [6]. Fig. 5 shows some examples of where the model fails and succeeds in predicting input handwritten images.

The HADARA80P dataset was mainly used in word spotting systems thus, the reported results are at the word-level. In [23] the authors used the line-level HADARA80P dataset with other non-historical Arabic datasets to train their system. They reported the overall accuracy of the system using all datasets. This makes our research a pilot study for this dataset.

TABLE 7. COMPARISON OF THE CER AND WER (%) ACHIEVED IN THIS WORK WITH THE PREVIOUSLY REPORTED COMPETITIVE STATE-OF-THE-ART RESULTS ON THE IAM DATASET.

| System | Method | CER | WER |
|---|---|---|---|
| Chenetal.[3] | CNN&LSTM | 11.15 | 34.55 |
| Phametal.[4] | CNN&LSTM | 10.8 | 35.1 |
| Khrishnanetal.[24] | CNN | 9.78 | 32.89 |
| Chowdhuryetal.[25] | CNN&RNN | 8.10 | 16.70 |
| Puigcerver[1] | CNN&LSTM | 6.2 | 20.2 |
| Markouetal.[26] | CNN&LSTM | 6.14 | 20.04 |
| Duttaetal.[27] | CNN&LSTM | 5.8 | 17.8 |
| Tassopoulouetal.[28] | CNN&LSTM | 5.18 | 17.68 |
| Yousefetal.[29] | CNN | 4.9 | - |
| Michaeletal.[2] | CNN&LSTM | 4.87 | - |
| Cojocaruetal.[30] | CNN&DeformableCNN | 4.6 | 19.3 |
| Retsinasetal.[6] | CNN&DeformableCNN | 4.67 | 16.57 |
| **Proposed** | **CNN&DeformableCNN&Self-ONNs** | **4.53** | **15.08** |

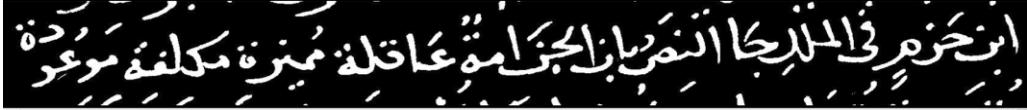

Actual Text: ابن حزم فى الملل جا النص بان الجن امة عاقلة مميزة مكلفة موعودة

Predicted Text: ابن حزم فى المـالـال جا النص بان الجن امة اعاقلة ماميزة مكلفة موعودا

CER: 5.0

WER: 4.0

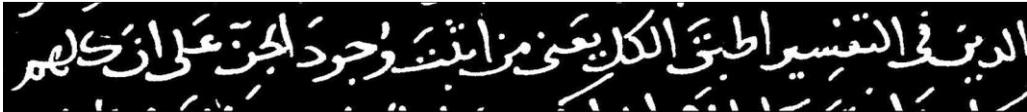

Actual Text: الدين فى التفسير اطبق الكل يعنى من اثبت وجود الجن على ان كلهم

Predicted Text: الدين فى التفسير اطبق الكل يعنى امن اثبت وجود الجن اعلى ان كلهما

CER: 3.0

WER: 3.0

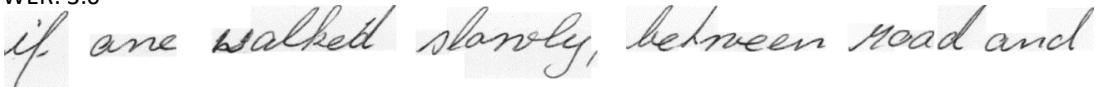

Actual Text: if one walked slowly, between road and
Predicted Text: if are walked slanely, between Noad and
CER:6.0
WER:3.0

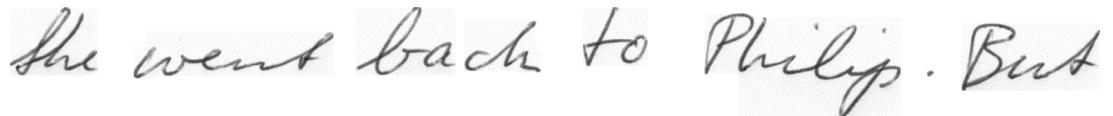

Actual Text: She went back to Philip. But
Predicted Text: the went back to Philigs. But
CER:3.0

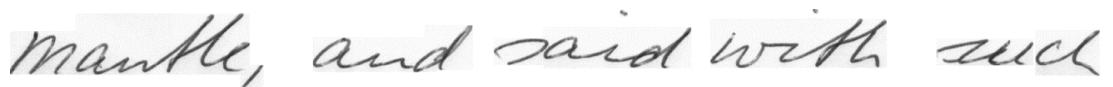

Actual Text: mantle, and said with such
Predicted Text: mantle, and said with such
CER:0.0
WER:0.0

Fig. 5. Sample of Model predictions of the HADARA80 P and the IAM datasets with their CER and WER.

## 5 CONCLUSION

In this paper, a novel approach based on Self-ONNs is proposed for HTR. Reaching the state-of-the-art performance levels in the IAM English dataset and superior performance in the HADARA80P Arabic dataset, the proposed approach benefits from the superior learning capabilities of the Self-ONNs that are heterogeneous network models with generative neurons. The previous top model proposed in [6] employs an uncertainty reduction method to improve the overall accuracy to 4.55% CER and 16.08% WER on the IAM line dataset. The proposed Self-ONN-based approach surpasses the original model even without employing any uncertainty reduction or any other post-processing whilst the network depth is further reduced. The exceptional margin between the results of the CNN model versus the Self-ONN model in the noisy HADARA80P dataset confirms the superior learning capabilities of Self-ONNs in such old and degraded manuscripts. This study shows that the 1D operational layers with generative neurons are able to represent complicated contextual information and handle HTR efficiently. Accuracy performance monitoring shows that, in both datasets, operational layers is the layers after the input layer of the model which indicates the importance of the low-level features in the HTR task. The optimized PyTorch implementation of Self-ONNs is publicly shared in [31].

The future work of this research will explore different training strategies for Self-ONNs and investigate the performance of Self-ONNs against other types of deep structures like transformers [32]–[34], tensor-based learning [35], and Convolution Long-Short Term Memory (CLSTM) [36]. Moreover, an investigation of the use of document summarization techniques [37] in the context of document image processing will also be considered.

## REFERENCES

[1] J. Puigcerver, "Are Multidimensional Recurrent Layers Really Necessary for Handwritten Text Recognition?," in *2017 14th IAPR International Conference on Document Analysis and Recognition (ICDAR)*, Nov. 2017, vol. 1, pp. 67–72. doi: 10.1109/ICDAR.2017.20.

[2] J. Michael, R. Labahn, T. Gruning, and J. Zollner, "Evaluating Sequence-to-Sequence Models for Handwritten Text Recognition," in *2019 International Conference on Document Analysis and Recognition (ICDAR)*, Sep. 2019, pp. 1286–1293. doi: 10.1109/ICDAR.2019.00208.

[3] Z. Chen, Y. Wu, F. Yin, and C.-L. Liu, "Simultaneous Script Identification and Handwriting Recognition via Multi-Task Learning of Recurrent Neural Networks," in *2017 14th IAPR International Conference on Document Analysis and Recognition (ICDAR)*, Nov. 2017, vol. 1, pp. 525–530. doi: 10.1109/ICDAR.2017.92.

[4] V. Pham, T. Bluche, C. Kermorvant, and J. Louradour, "Dropout Improves Recurrent Neural Networks for Handwriting Recognition," in *2014 14th International Conference on Frontiers in Handwriting Recognition*, Sep. 2014, pp. 285–290. doi: 10.1109/ICFHR.2014.55.

[5] D. Castro, B. L. D. Bezerra, and M. Valenca, "Boosting the Deep Multidimensional Long-Short-Term Memory Network for Handwritten Recognition Systems," in *2018 16th International Conference on Frontiers in Handwriting Recognition (ICFHR)*, Aug. 2018, pp. 127–132. doi: 10.1109/ICFHR-2018.2018.00031.

[6] G. Retsinas, G. Sfikas, C. Nikou, and P. Maragos, "Deformation-Invariant Networks For Handwritten Text Recognition," in *2021 IEEE International Conference on Image Processing (ICIP)*, Sep. 2021, pp. 949–953. doi: 10.1109/ICIP42928.2021.9506414.

[7] J. Dai *et al.*, "Deformable Convolutional Networks," in *2017 IEEE International Conference on Computer Vision (ICCV)*, Oct. 2017, pp. 764–773. doi: 10.1109/ICCV.2017.89.

[8] M. Gabbouj *et al.*, "Robust Peak Detection for Holter ECGs by Self-Organized Operational Neural Networks," *IEEE Transactions on Neural Networks and Learning Systems*, pp. 1–12, 2022, doi: 10.1109/TNNLS.2022.3158867.

[9] J. Malik, O. C. Devecioglu, S. Kiranyaz, T. Ince, and M. Gabbouj, "Real-Time Patient-Specific ECG Classification by 1D Self-Operational Neural Networks," *IEEE Transactions on Biomedical Engineering*, pp. 1–1, 2021, doi: 10.1109/TBME.2021.3135622.

[10] J. Malik, S. Kiranyaz, and M. Gabbouj, "Operational vs Convolutional Neural Networks for

[11]  J. Malik, S. Kiranyaz, and M. Gabbouj, "Self-organized operational neural networks for severe image restoration problems," *Neural Networks*, vol. 135, pp. 201–211, Mar. 2021, doi: 10.1016/j.neunet.2020.12.014.

[12]  S. Kiranyaz, T. Ince, A. Iosifidis, and M. Gabbouj, "Operational neural networks," *Neural Computing and Applications*, vol. 32, no. 11, pp. 6645–6668, Jun. 2020, doi: 10.1007/s00521-020-04780-3.

[13]  D. T. Tran, S. Kiranyaz, M. Gabbouj, and A. Iosifidis, "Heterogeneous Multilayer Generalized Operational Perceptron," *IEEE Transactions on Neural Networks and Learning Systems*, vol. 31, no. 3, pp. 710–724, Mar. 2020, doi: 10.1109/TNNLS.2019.2914082.

[14]  S. Kiranyaz, T. Ince, A. Iosifidis, and M. Gabbouj, "Generalized model of biological neural networks: Progressive operational perceptrons," in *2017 International Joint Conference on Neural Networks (IJCNN)*, May 2017, pp. 2477–2485. doi: 10.1109/IJCNN.2017.7966157.

[15]  S. Kiranyaz, J. Malik, H. ben Abdallah, T. Ince, A. Iosifidis, and M. Gabbouj, "Self-organized Operational Neural Networks with Generative Neurons," *Neural Networks*, vol. 140, pp. 294–308, Aug. 2021, doi: 10.1016/j.neunet.2021.02.028.

[16]  A. Rahman *et al.*, "Robust biometric system using session invariant multimodal EEG and keystroke dynamics by the ensemble of self-ONNs," *Computers in Biology and Medicine*, vol. 142, p. 105238, Mar. 2022, doi: 10.1016/j.compbiomed.2022.105238.

[17]  M. Soltanian, J. Malik, J. Raitoharju, A. Iosifidis, S. Kiranyaz, and M. Gabbouj, "Speech Command Recognition in Computationally Constrained Environments with a Quadratic Self-Organized Operational Layer," in *2021 International Joint Conference on Neural Networks (IJCNN)*, Jul. 2021, pp. 1–6. doi: 10.1109/IJCNN52387.2021.9534232.

[18]  K. He, X. Zhang, S. Ren, and J. Sun, "Deep Residual Learning for Image Recognition," in *2016 IEEE Conference on Computer Vision and Pattern Recognition (CVPR)*, Jun. 2016, pp. 770–778. doi: 10.1109/CVPR.2016.90.

[19]  A. Graves, S. Fernández, F. Gomez, and J. Schmidhuber, "Connectionist temporal classification," in *Proceedings of the 23rd international conference on Machine learning - ICML '06*, 2006, pp. 369–376. doi: 10.1145/1143844.1143891.

[20]  U.-V. Marti and H. Bunke, "The IAM-database: an English sentence database for offline handwriting recognition," *International Journal on Document Analysis and Recognition*, vol. 5, no. 1, pp. 39–46, Nov. 2002, doi: 10.1007/s100320200071.

[21]  W. Pantke, M. Dennhardt, D. Fecker, V. Margner, and T. Fingscheidt, "An Historical Handwritten Arabic Dataset for Segmentation-Free Word Spotting - HADARA80P," in *2014 14th International Conference on Frontiers in Handwriting Recognition*, Sep. 2014, vol. 2014-Decem, pp. 15–20. doi: 10.1109/ICFHR.2014.11.

[22]  V. I. Levenshtein and others, "Binary codes capable of correcting deletions, insertions, and reversals," in *Soviet physics doklady*, 1966, vol. 10, no. 8, pp. 707–710. doi: 1966SPhD...10..707L.

[23]  F. Stahlberg and S. Vogel, "QATIP - An Optical Character Recognition System for Arabic Heritage Collections in Libraries," 2016. doi: 10.1109/DAS.2016.81.

[24]  P. Krishnan, K. Dutta, and C. V. Jawahar, "Word Spotting and Recognition Using Deep Embedding," in *2018 13th IAPR International Workshop on Document Analysis Systems (DAS)*, Apr. 2018, pp. 1–6. doi: 10.1109/DAS.2018.70.

[25]  A. Chowdhury and L. Vig, "An efficient end-to-end neural model for handwritten text recognition," *arXiv preprint arXiv:1807.07965*, 2018.

[26]  K. Markou *et al.*, "A Convolutional Recurrent Neural Network for the Handwritten Text Recognition of Historical Greek Manuscripts," in *International Conference on Pattern Recognition*, 2021, pp. 249–262. doi: 10.1007/978-3-030-68787-8_18.

[27]  K. Dutta, P. Krishnan, M. Mathew, and C. V. Jawahar, "Improving CNN-RNN Hybrid Networks for Handwriting Recognition," in *2018 16th International Conference on Frontiers in Handwriting Recognition (ICFHR)*, Aug. 2018, pp. 80–85. doi: 10.1109/ICFHR-2018.2018.00023.

[28]  V. Tassopoulou, G. Retsinas, and P. Maragos, "Enhancing Handwritten Text Recognition with N-gram sequence decomposition and Multitask Learning," in *2020 25th International Conference on Pattern Recognition (ICPR)*, 2021, pp. 10555–10560. doi: 10.48550/arXiv.2012.14459.

[29] M. Yousef, K. F. Hussain, and U. S. Mohammed, "Accurate, data-efficient, unconstrained text recognition with convolutional neural networks," *Pattern Recognition*, vol. 108, p. 107482, Dec. 2020, doi: 10.1016/j.patcog.2020.107482.

[30] I. Cojocaru, S. Cascianelli, L. Baraldi, M. Corsini, and R. Cucchiara, "Watch Your Strokes: Improving Handwritten Text Recognition with Deformable Convolutions," in *2020 25th International Conference on Pattern Recognition (ICPR)*, Jan. 2021, pp. 6096–6103. doi: 10.1109/ICPR48806.2021.9412392.

[31] Self-ONNs, "http://selfonn.net/," 2021.

[32] M. Kaselimi, A. Voulodimos, I. Daskalopoulos, N. Doulamis, and A. Doulamis, "A Vision Transformer Model for Convolution-Free Multilabel Classification of Satellite Imagery in Deforestation Monitoring," *IEEE Transactions on Neural Networks and Learning Systems*, 2022, doi: 10.1109/TNNLS.2022.3144791.

[33] M. Raghu, T. Unterthiner, S. Kornblith, C. Zhang, and A. Dosovitskiy, "Do Vision Transformers See Like Convolutional Neural Networks?," *Advances in Neural Information Processing Systems*, vol. 34, 2021, doi: 10.48550/arXiv.2108.08810.

[34] W. Liu *et al.*, "Is the aspect ratio of cells important in deep learning? A robust comparison of deep learning methods for multi-scale cytopathology cell image classification: From convolutional neural networks to visual transformers," *Comput Biol Med*, vol. 141, p. 105026, 2022, doi: 10.48550/arXiv.2105.07402.

[35] K. Makantasis, A. D. Doulamis, N. D. Doulamis, and A. Nikitakis, "Tensor-based classification models for hyperspectral data analysis," *IEEE Transactions on Geoscience and Remote Sensing*, vol. 56, no. 12, pp. 6884–6898, 2018, doi: 10.1109/TGRS.2018.2845450.

[36] A. Sarabu and A. K. Santra, "Human action recognition in videos using convolution long short-term memory network with spatio-temporal networks," *Emerging Science Journal*, vol. 5, no. 1, pp. 25–33, 2021, doi: 10.28991/esj-2021-01254.

[37] K. K. Mamidala and S. K. Sanampudi, "A novel framework for multi-document temporal summarization (Mdts)," *Emerging Science Journal*, vol. 5, no. 2, pp. 184–190, 2021, doi: 10.28991/esj-2021-01268.